%% file: predictive2014_arxiv.tex
\documentclass{article}
\input{preamble_arxiv.tex}

\begin{document} 

\twocolumn[
\icmltitle{Modeling sequential data using \\higher-order relational features and predictive training}

\icmlauthor{Vincent Michalski}{vmichals@rz.uni-frankfurt.de}
\icmladdress{Goethe-University Frankfurt,
             Frankfurt, Germany}
\icmlauthor{Roland Memisevic}{roland.memisevic@umontreal.ca}
\icmladdress{University of Montreal,
            Montreal, Canada}
\icmlauthor{Kishore Konda}{konda@informatik.uni-frankfurt.de}
\icmladdress{Goethe-University Frankfurt,
             Frankfurt, Germany}

\icmlkeywords{higher-order relational features, predictive training, motion}

\vskip 0.3in
]

\begin{abstract} 
    \input{abstract}

\end{abstract} 

\section{Introduction}
\label{sec:introduction}
\input{intro}

\section{Relational feature learning}

\label{sec:gae}
\input{gae}

\section{Higher-order relational features}
\subsection{Approximating discrete-time dynamical systems}
\label{sec:approx_dyn_sys}
\input{approx_dyn_sys}

\input{higherorder_features}
\subsection{The higher-order gated autoencoder}
\label{sec:hgae}
\input{hgae}

\section{Predictive training of relational models}
\subsection{Single-step prediction}
\label{sec:pred_training}
\input{pred_training}

\subsection{Multi-step prediction}
\input{learn_on_seq}

\section{Experiments}
\input{experiments.tex}

\subsection{Preprocessing and initialization}
\label{sec:preprocessing}
\input{preprocessing}
\subsection{Comparison of predictive and reconstructive training}
\input{pred_vs_rec}

\subsection{Detecting acceleration}
\label{sec:acc_transformations}
\input{acc_transformations}
\subsection{Sequence prediction}
\input{seq_pred}
\section{Discussion}
\input{discussion}

\section*{Acknowledgments} 
\input{acknowledgements}

\bibliography{predictive2014_arxiv}
\bibliographystyle{icml2014}

\end{document}

%% file: preamble_arxiv.tex
\usepackage{amsmath}
\usepackage{times}
\usepackage{graphicx}
\usepackage{subfigure} 
\usepackage{natbib}
\usepackage{algorithm}
\usepackage{algorithmic}
\usepackage{hyperref}

\usepackage[accepted]{style} 

\icmltitlerunning{Modeling sequential data using
  higher-order relational features and predictive training
}
\usepackage{multirow}

\usepackage{ifthen}
\newboolean{showNotes}

\setboolean{showNotes}{true}
\newcommand{\todo}[1]{\ifthenelse{\boolean{showNotes}}{\textcolor{red}{\textbf{\textcolor{red}{(TODO: #1)}}}}{}}

\newcommand{\m}[2][\empty]{
    \ifthenelse{\equal{#1}{\empty}}
    {\ifthenelse{\equal{#2}{\empty}}
        {\mathbf{m}}
        {
            {\mathbf{m}^{(#2)}}
        }
    }
    {
        {\mathbf{m}_{#1}^{(#2)}}
    }
}
\newcommand{\predm}[2][\empty]{
    \ifthenelse{\equal{#1}{\empty}}
    {\ifthenelse{\equal{#2}{\empty}}
        {\hat{\mathbf{m}}}
        {
            {\hat{{\mathbf{m}}^{(#2)}}}
        }
    }
    {
        {\hat{\mathbf{m}}_{#1}^{(#2)}}
    }
}

\newcommand{\prnn}[2][\empty]{
    \ifthenelse{\equal{#1}{\empty}}
    {\mathbf{p}_{#2}}
    {\mathbf{p}_{#2}^{#1}}
}

\newcommand{\x}[1][]{\mathbf{x}^{(#1)}}
\newcommand{\predx}[1][]{\hat{\mathbf{x}}{}^{(#1)}}
\newcommand{\recx}[1][]{\tilde{\mathbf{x}}{}^{(#1)}}

\newcommand{\U}[1][\empty]{
    \ifthenelse{\equal{#1}{\empty}}
    {\mathbf{U}_{} }
    {\mathbf{U}_{#1}}
}
\newcommand{\V}[1][\empty]{
    \ifthenelse{\equal{#1}{\empty}}
    {\mathbf{V}{} }
    {\mathbf{V}_{#1}}
}

\newcommand{\W}[1][\empty]{
    \ifthenelse{\equal{#1}{\empty}}
    {\mathbf{W}{} }
    {\mathbf{W}_{#1}}
}

\newcommand{\bigrnd}[1]{\big( #1 \big)}

\newcommand{\T}[0]{\mathrm{T}}

%% file: abstract.tex
Bi-linear feature learning models, like the gated autoencoder, were proposed as a way 
to model relationships between frames in a video. By minimizing reconstruction error of 
one frame, given the previous frame, these models learn ``mapping units'' that encode 
the transformations inherent in a sequence, and thereby learn to encode motion.  
In this work we extend bi-linear models by introducing ``higher order mapping units'' 
that allow us to encode transformations between frames \emph{and} transformations 
between transformations. 

We show that this makes it possible to encode temporal structure that is more 
complex and lon\-ger-range than the structure captured within standard bi-linear models.  
We also show that a natural way to train the model is by replacing the commonly
used reconstruction objective with a \emph{prediction} objective which forces the 
model to correctly predict the evolution of the input multiple steps into the future.\\ 
Learning can be achieved by back-pro\-pa\-ga\-ting the mul\-ti-step prediction through time. 
We test the model on various temporal prediction tasks, and show that higher-order 
mappings and predictive training both yield a significant improvement over bi-linear models 
in terms of prediction accuracy.

%% file: intro.tex
We explore the application of relational feature learning models \citep[e.g.][]{imtrans,taylorFactoredConditional}
in sequence modeling. 
To this end, we propose a bilinear model to describe frame-to-frame transitions in 
image sequences. 
In contrast to existing work on modeling relations, we propose a new training scheme, 
which we call predictive training: 
after a transformation is extracted from two frames in the video, 
the model tries to predict the next frame by assuming constancy of the transformations 
through time. 

We then introduce a deep bilinear model as a natural application of predictive training, 
and as a way to relax the assumption of constancy of the transformation. 

The model learns relational features, as well as  ``higher-order relational features'', 
representing relations between the transformations themselves.
To this end, the bottom-layer bilinear model infers a representation of motion 
from two seed frames as well as a representation of motion from two later frames. 
The top layer is itself a bilinear model, that learns to represent the relation between 
the inferred lower-level transformations. It can be thought of as learning a 
second-order ``derivative'' of the temporal evolution of the high-dimensional input time series. 
We show that an effective way to train these models is to first pre-train the layers individually 
using pairs of frames for the bottom layer and pairs of inferred transformations for the next layer, 
and to subsequently fine-tune parameters using complete sequences, by back-propagating a multi-step 
lookahead cost through time. 

The model as a whole may be thought of as a way to model a dynamical system as a 
second order partial difference equation.  
While in principle the model could be stacked to take into account differences of arbitrary 
order, we demonstrate that the two-layer model is surprisingly effective at modeling a variety 
of complicated image sequences. 

Both layers of our model make use of multiplicative interactions between filter responses in order 
to model relations \cite{memisevic2013learning}.
Multiplicative interactions 
were recently shown to be useful in recurrent neural networks by \cite{sutskever2011generating}. 
In contrast to our work, \cite{sutskever2011generating} use multiplicative interactions to gate 
the connections between hidden states, so that each observation can be thought of as blending 
in a separate hidden state transition. A natural application of this is sequences of discrete symbols,
and the model is consequently demonstrated on text. 
In our work, the role of multiplicative interactions is explicitly to yield encodings 
of transformations, such as frames in a video, and we apply the model primarily to video data. 

Our model also bears some similarity to \cite{taylorFactoredConditional} who model MOCAP data 
using a generatively trained three-way Restricted Boltzmann Machine, where a second layer of hidden 
units can be used to model more ``abstract'' features of the time series. 
In contrast to that work, our higher-order units which are three-way units too, 
are used to expressly model higher-order transformations
(transformations between the transformations learned in the
first layer).
Furthermore, we show that predictive fine-tuning using backprop through time allows us to train 
the model discriminatively and yields much better performance than generative training by itself.

%% file: gae.tex
\begin{figure}[ht]
\vskip 0.2in
\begin{center}
\centerline{\includegraphics[width=0.6\columnwidth]{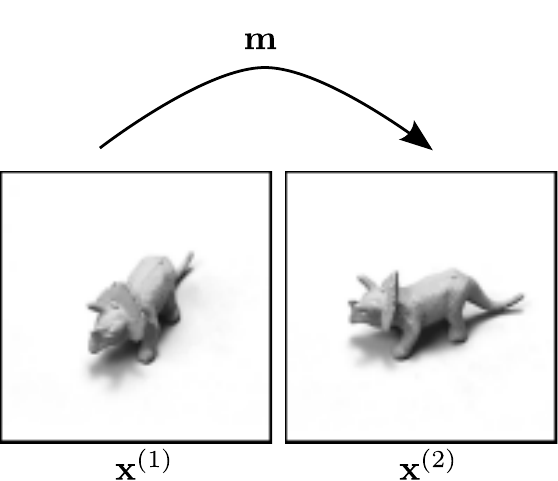}}
\caption{The relational features $\mathbf{m}$ represent the correspondences between
two inputs $\x[1]$ and $\x[2]$.}
\label{fig:rel_feat}
\end{center}
\vskip -0.2in
\end{figure}

In order to learn features, $\mathbf{m}$, that represent the relationship between two frames 
$\x[1]$ and $\x[2]$ as shown in Figure \ref{fig:rel_feat},  
it is necessary to learn a basis that can represent the correlation structure across 
the frames. 

In a video, given one frame $\x[1]$ there can be a multitude of potential next frames 
$\x[2]$. It is therefore common to use bi-linear models, like the 
Gated Boltzmann Machine (GBM) \cite{taylorFactoredConditional}, 
the Gated Autoencoder (GAE) \cite{memisevic2011gradient}, 
and similar models \citep[see][for an overview]{memisevic2013learning}
whose hidden variables can represent which transformation, out of the pool 
of many possible transformations, can take $\x[1]$ to $\x[2]$. 

More formally, bi-linear models learn to represent the linear 
transformation, $\mathbf{L}$,
between two images $\x[1]$ and $\x[2]$, where 
\begin{equation}
\x[2]=\mathbf{L}\x[1]
\end{equation}
It can be shown that a weighted sum of \emph{products of filter responses} 
is able to identify the transformation. The reason is that the weighted 
sum is large if the angle between filters is similar to the angle (in ``pixel-space'')
between the two frames. 
That way, hidden units represent the observed transformation in the form of 
a set of phase-differences in the invariant subspaces of the transformation 
class \cite{memisevic2013learning}.
As hidden units encode the transformation between images, rather than the content of 
the images, they are commonly referred to as \emph{mapping units}.
We shall focus on the autoencoder variant of these models for the purposes of 
this paper, but one can use other models such as the GBM.

\begin{figure}[ht]
\vskip 0.2in
\begin{center}
\centerline{\includegraphics[width=0.95\columnwidth]{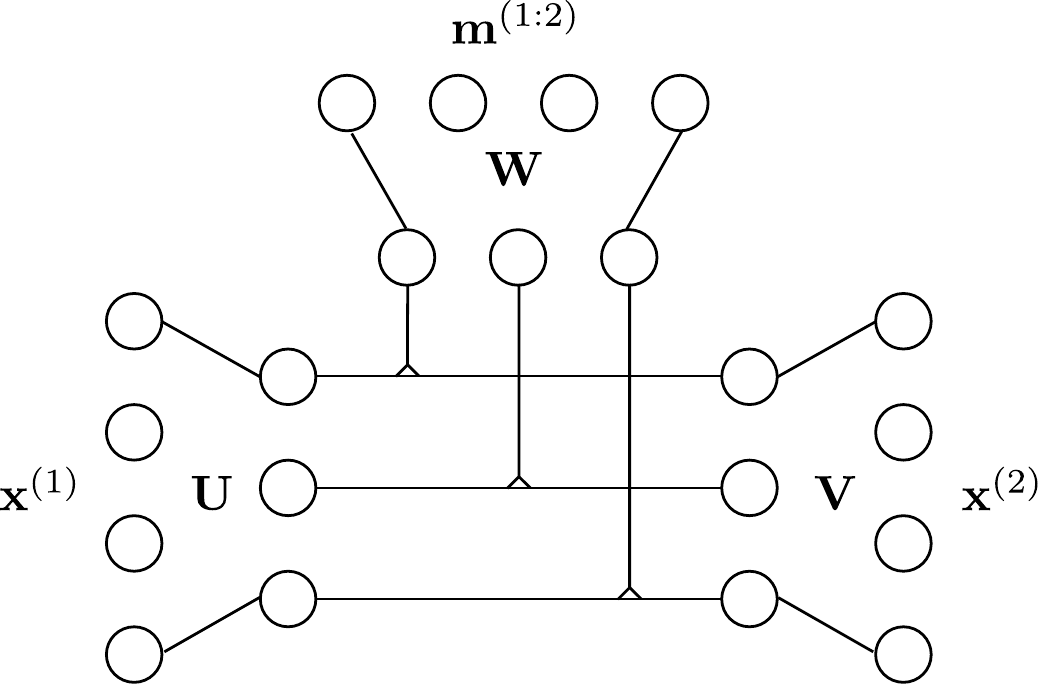}}
\caption{Graphical representation of the gated autoencoder. The two inputs
$\x[1]$ and $\x[2]$ are projected onto features and the mapping units pool
over pairwise products of these features.}
\label{fig:gae}
\end{center}
\vskip -0.2in
\end{figure}

Formally, the response of a mapping unit layer can be written 
\begin{equation}
    \mathbf{m} = \sigma\big(\W(\U\x[1] \odot \V\x[2])\big)
    \label{eq:mappings}
\end{equation}
where $\U, \V$ and $\W$ are parameter matrices, 
and where 
$\odot$ denotes elementwise multiplication. Further, 
$\sigma$ is an elementwise non-linearity, such as the logistic sigmoid.  

Given mapping unit activations, $\mathbf{m}$, 
as well as the first image, 
the second image can be reconstructed by applying the transformation encoded in $\mathbf{m}$ 
as follows \cite{memisevic2011gradient}:
\begin{equation}
    \recx[2] = \V^\T
    \bigrnd{
        \U\x[1] \odot \W^\T\m{}
    }.
    \label{eq:gae_rec2}
\end{equation}
As the model is symmetric, we can likewise define the reconstruction of the first image 
given the second as:
\begin{equation}
    \recx[1] = \U^\T
    \bigrnd{
        \V\x[2] \odot \W^\T\m{}
    }
    \label{eq:gae_rec1}
\end{equation}
from which one obtains the reconstruction error
\begin{equation}
    \mathcal{L} = 
    ||\x[1]-\recx[1]||^2 +
    ||\x[2]-\recx[2]||^2
    \label{eq:gae_rec_error}
\end{equation}
for training. 
It can be shown that minimizing reconstruction error on image pairs 
will turn each row in $\U$ and the corresponding row in $\V$ into a pair 
of phase-shifted filters. Together the filters span the invariant subspaces
of the transformation class inherent in the training pairs with which 
the model was trained. 
As as result, each component of $\mathbf{m}$ is tuned to a phase-delta after learning,
and it is independent of the absolute phase of each image \cite{memisevic2013learning}.

%% file: approx_dyn_sys.tex
A quite natural extension of the concept of relational features can be motivated
by looking at relational models as performing a kind of first-order Taylor 
approximation of the input sequence, where 
the hidden representation models the partial first-order derivatives of the inputs 
with respect to time.
Based on this view, we propose an approach that exploits correlations between subsequent 
sequence elements to model a dynamical system which approximates the sequence.
This is a very different way to address long-range correlations than assuming 
memory units that explicitly keep state \cite{hochreiter1997long}. 
Instead, here we assume that there is structure in the temporal evolution of the 
input stream and we focus on capturing this structure.

As an intuitive example, consider a video that is known to be a sinusoidal signal, 
but with unkown frequency, phase and motion direction. 
The complete video can be specified exactly and completely by the first three 
seed images. Therefore, given these three images, we would in principle be able 
to predict the rest of the video ad infinitum.

The first-order partial derivative of a multidimensional dis\-crete-time dynamical system 
describes the correspondences between state vectors at subsequent time steps.
Relational feature learning applied to subsequent elements of a sequence 
can be viewed as a way to learn these derivatives, suggesting 
that we may model higher-order partial derivatives with higher-order relational features.

%% file: higherorder_features.tex
\begin{figure}[ht]
\vskip 0.2in
\begin{center}
\centerline{\includegraphics[width=.85\columnwidth]{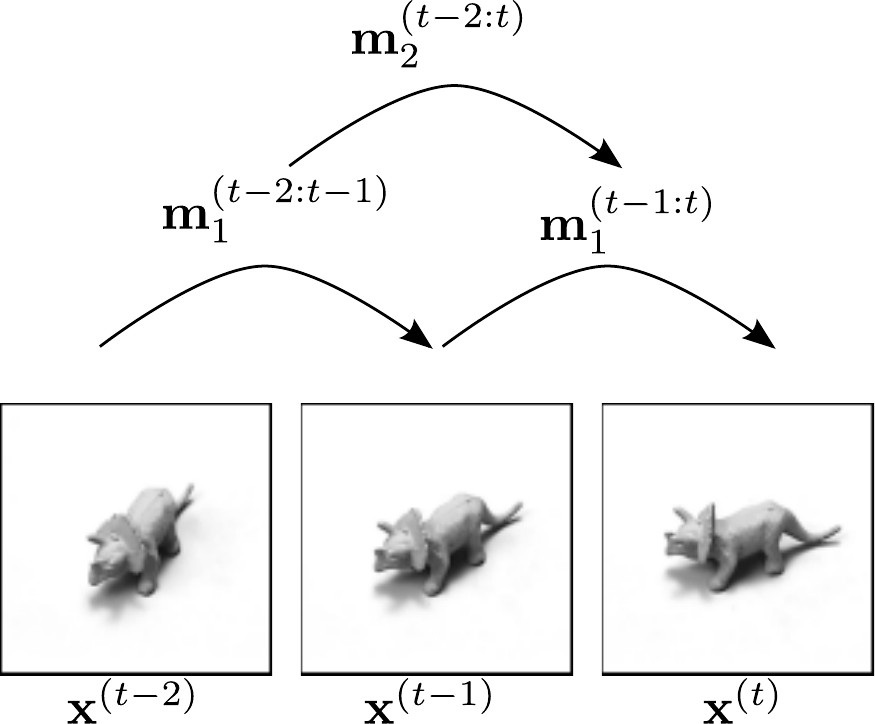}}
\caption{First-order relational features $\m[1]{t-2:t-1}$ and $\m[1]{t-1:t}$ describe correspondences 
between multiple entities, e.g. two frames of a video. The second-order relational features 
$\m[2]{t-2:t}$ describe correspondences between the first-order relational features.}
\label{fig:higher_order_relational_feature}
\end{center}
\vskip -0.2in
\end{figure}

We model second-order derivatives 
by cascading relational features in a pyramid as 
depicted\footnote{Images taken 
from the NORB data set described in \cite{lecun2004learning}} in 
\ref{fig:higher_order_relational_feature}.
Given a sequence of inputs $\x[t-2],\x[t-1],\x[t]$, first-order relational 
features $\m[1]{t-1:t}$ describe the transformations between two subsequent 
inputs $\x[t-1]$ and $\x[t]$. Second-order relational features $\m[2]{t-2:t}$
describe correspondences between two first-order relational features
$\m[1]{t-2:t-1}$ and $\m[1]{t-1:t}$, modeling the analog of the partial second-order 
derivatives of the inputs with respect to time.
Section \ref{sec:acc_transformations} presents experiments with two layers of relational 
features that support this view. 

%% file: hgae.tex
We implement a higher-order gated autoencoder (HGAE) using the following  
modular approach. 
The second-order HGAE is constructed using two GAE modules, one that
relates inputs and another that relates mappings of the first GAE.

The first-layer GAE instance models correspondences between input pairs 
using filter matrices $\U[1],\V[1]$ and $\W[1]$ (the subscript index refers to 
the layer).
Using the first-layer GAE, mappings $\m[1]{t-2:t-1}$ and $\m[1]{t-1:t}$ for overlapping 
input pairs $(\x[t-2],\x[t-1])$ and $(\x[t-1],\x[t])$ are inferred and 
this pair of first-layer mappings is used as input for a second GAE instance.
This second GAE models correspondences between the mappings of the first-layer using
filter matrices $\U[2],\V[2]$ and $\W[2]$. 

For the two-layer model, inference amounts to computing first- and second-order 
mappings according to 
\begin{align}
  \m[1]{t-2:t-1} = \sigma\big(
    \W[1] \big( 
      &(\U[1] \x[t-2]) \odot (\V[1] \x[t-1])
   \big)\big)\\
   \m[1]{t-1:t} = \sigma\big(
    \W[1] \big( 
      &(\U[1] \x[t-1]) \odot (\V[1] \x[t])
   \big)\big)
  \label{eq:hgae_layer1}\\
  \m[2]{t-2:t} = \sigma\big(
    \W[2] \big( 
      &(\U[2] \m[1]{t-2:t-1}) \odot 
      (\V[1] \m[1]{t-1:t})
   \big)\big)
   \label{eq:hgae_layer2}
\end{align}

Cascading GAE modules in this way can also be motivated from the perspective
of orthogonal transformations as subspace rotations. As stated in \cite{memisevic2013learning},
summing over filter-response products can yield transformation detectors which 
are invariant to the initial phase of the transformation and also partially invariant
to the content of the images. 
The relative rotation angle (phase delta) between two projections is itself an angle, and 
their relation can be viewed as an ``angular acceleration''.  

In contrast to the standard two-frame model, in this model reconstruction error is not 
directly applicable (although a naive way to train the model is to minimize 
reconstruction error for each pair of adjacent nodes in each layer). 
However, there is a more natural way to train the model if training data forms a sequence, 
as we discuss next.

%% file: pred_training.tex
Given the first two frames of a sequence $\x[1], \x[2], \x[3]$ one can use 
the GAE to compute a prediction of the third frame as follows. 
First, mappings $\m{1,2}$ are inferred from $\x[1]$ and $\x[2]$ (see Equation \ref{eq:mappings})
and then used to compute a prediction $\predx[3]$ by applying the inferred transformation 
$\m{1,2}$ 
to frame $\x[2]$. 
Applying the transformation amounts to computing: 
\begin{equation}
 \predx[3] = \V^\T
 \big(
   \U \x[2] \odot \W^\T\m{1,2}
 \big)
 \label{eq:pred_train}
\end{equation}
This prediction of $\x[3]$ will be a good prediction under the assumption that the frame-to-frame transformations 
from $\x[1]$ to $\x[2]$ and from $\x[2]$ to $\x[3]$ are approximately the same, in other words if transformations 
themselves are assumed to be approximately \emph{constant} in time. 

In this case, one can train the GAE to minimize the prediction error  
\begin{equation}
 \mathcal{L} = ||\predx[3] - \x[3]||_{2}^2
 \label{eq:pred_error}
\end{equation}
instead of minimizing the reconstruction error in Equation \ref{eq:gae_rec_error}.
This type of supervised training objective, in contrast to the standard GAE objective, can also 
guide the mapping representation to be invariant to image content, because encoding the content 
of $\x[2]$ will not in general help predicting $\x[3]$.

\begin{figure}[ht]
\vskip 0.2in
\begin{center}
\centerline{\includegraphics[width=.85\columnwidth]{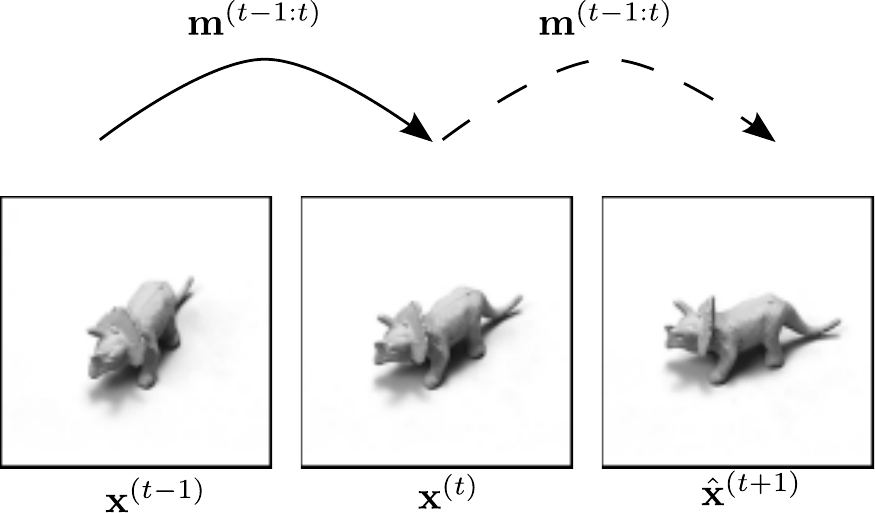}}
\caption{The assumption of similarity between the transformations from $\x[t-1]$ to $\x[t]$ and 
from $\x[t]$ to $\x[t+1]$ allows us to define a prediction $\predx[t+1]$ by applying the 
inferred transformation $\m{t-1:t}$ to $\x[t]$.}
\label{fig:gae_pred}
\end{center}
\vskip -0.2in
\end{figure}

When the assumption of constancy of the transformations is violated, we can use a higher layer to 
model how transformations themselves change over time. 
This will require a farther look-ahead for predictive training 
which we discuss in the following.

%% file: learn_on_seq.tex
\label{sec:learnonseq}
One can iterate the inference-prediction process to look more than one frame ahead in time.
To compute a prediction $\predx[4]$ one infers mappings from $\x[2]$ and $\predx[3]$:
\begin{equation}
 \m{2:3} = \sigma\bigrnd{\W(\U\x[2] \odot \V\predx[3])}
\end{equation}
and computes the prediction
\begin{equation}
 \predx[4] = \V^\T
 \big(
   \U \x[3] \odot \W^\T\m{2:3}
 \big).
 \label{eq:pred_train_step2}
\end{equation}
Then mappings can be inferred again from $\predx[3]$ and $\predx[4]$ to compute a prediction of 
$\predx[5]$, and so on.

For the two-layer HGAE this amounts to the assumption that the second-order 
relational structure in the sequence changes slowly over time and 
under this assumption we compute a prediction $\predx[t+1]$ in two steps:
First a prediction is made of the first-order relational features describing the correspondence between
$\x[t]$ and $\x[t+1]$:
\begin{equation}
    \predm[1]{t:t+1} = {\V[2]}^\T
    \big(
        \U[2] \m[1]{t-1:t} \odot \W[2]^\T \m[2]{t-2:t}
    \big)
    \label{eq:hgae_predmap}
\end{equation}
Using this prediction of the transformation between $\x[t]$ and $\x[t+1]$ the 
prediction $\predx[t+1]$ is made as follows:
\begin{equation}
    \predx[t+1] = {\V[1]}^\T
    \big(
        \U[1] \x[t] \odot \W[1]^\T \predm[1]{t:t+1}
    \big)
    \label{eq:hgae_pred}
\end{equation}

\begin{figure}[ht]
\vskip 0.2in
\begin{center}
\centerline{\includegraphics[width=.95\columnwidth]{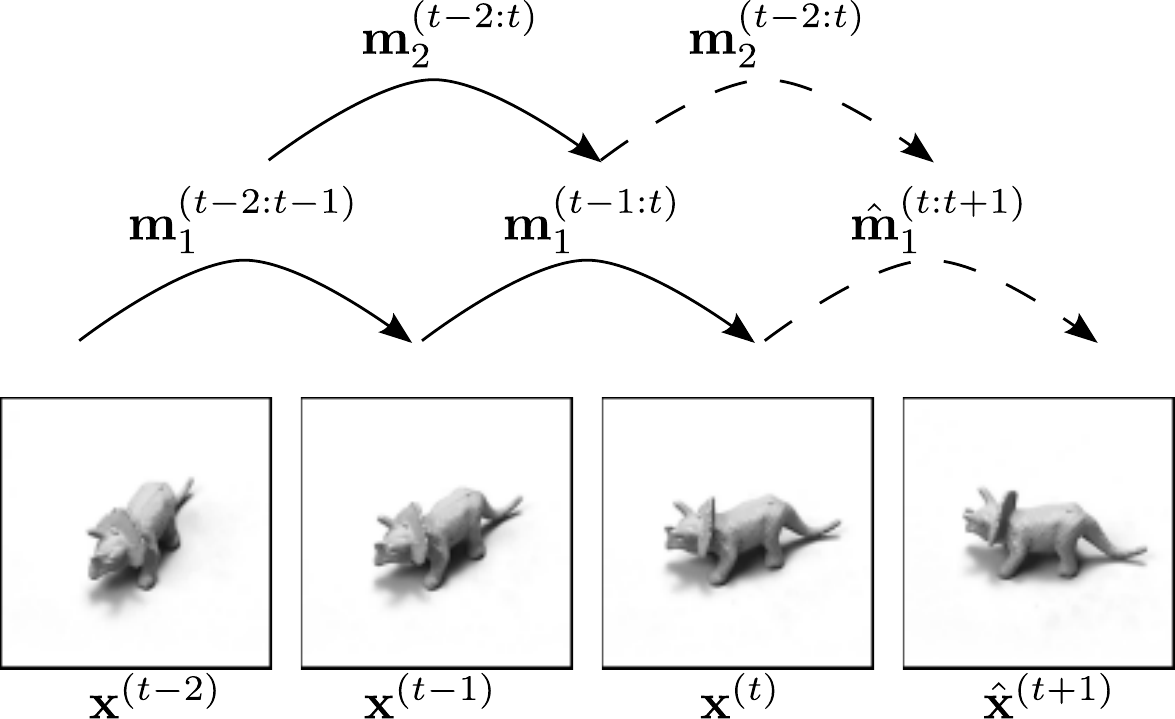}}
\caption{A prediction is made in two steps (the dashed lines) from top-to-bottom. The second-order
relational feature $\m[2]{t-2:t}$, inferred on the sequence $\x[t-2],\x[t-1],\x[t]$ is assumed
to be slowly changing and used to make a prediction of the first-order relational feature which
describes the correspondences between $\x[t]$ and $\x[t+1]$. This prediction is then used to
transform $\x[t]$ into the prediction $\predx[t+1]$.}
\label{fig:higher_order_relational_feature_pred}
\end{center}
\vskip -0.2in
\end{figure}

As with the GAE, one can predict multiple steps ahead in time using the HGAE by repeating the 
inference-prediction process 
on $\x[t-1], \x[t]$ and $\predx[t+1]$, i.e. by appending the prediction to the sequence and 
increasing $t$ by one.

The prediction process simply consists of iteratively computing predictions 
of the next lower level's activations beginning from the top.
To compute the top-level activations themselves, one needs a number of seed frames 
corresponding to the depth of the model. While two frames are sufficient to infer the 
transformations in the case of the GAE, three frames are required 
in the case of the two-layer model. 

The models can be trained using backprop through time \cite{werbos1988generalization}
to compute the gradients of the $k$-step ahead prediction error w.r.t. the parameters:
\begin{equation}
 \mathcal{L} = \sum\limits_{i=1}^k ||\predx[t+i] - \x[t+i]||_{2}^2
 \label{eq:kstep_pred_error}
\end{equation}

In our experiments, we observed that starting with single-step prediction, training and 
iteratively increasing the number of prediction steps during training considerably 
stabilizes the dynamics of the model and helps to prevent explosions in the magnitude of
the predictions.

%% file: experiments.tex
We tested and compared the models on videos with varying degrees of complexity, from synthetic constant to synthetic 
accelerated transformations to more complex real-world transformations.

%% file: preprocessing.tex
For all data sets PCA whitening was used for dimensionality reduction, 
retaining around $95\%$ of the variance.

Predictive training of the HGAE only worked after layerwise pre-training.
We used gradient descent with a learning rate of $0.001$ and momentum $0.9$.
Without pretraining the parameters did not converge to a useful configuration.
The first-layer GAE was trained to reconstruct pairs of subsequent sequence 
elements (as described in Section \ref{sec:gae}). Then pairs of mappings were 
computed on three subsequent inputs 
using the pretrained first-layer GAE. These mapping pairs were then used for
reconstructive pretraining of the second-layer GAE.

%% file: pred_vs_rec.tex
\begin{figure*}
    \centering
    \subfigure[\textsc{AccRot}]{
        \includegraphics[width=0.19\textwidth]{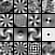}
        \hspace{0.02\textwidth}
        \includegraphics[width=0.19\textwidth]{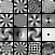}
    }
    \hspace{0.05\textwidth}
    \subfigure[\textsc{AccShift}] {
        \includegraphics[width=0.19\textwidth]{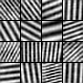}
        \hspace{0.02\textwidth}
        \includegraphics[width=0.19\textwidth]{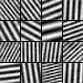}
    }
    \\

    \subfigure[Bouncing Balls] {
        \includegraphics[width=0.19\textwidth]{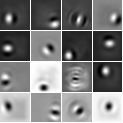}
        \hspace{0.02\textwidth}
        \includegraphics[width=0.19\textwidth]{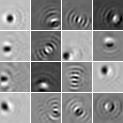}
        \label{fig:bb_filters}
    }
    \hspace{0.05\textwidth}
    \subfigure[NORBVideos] {
        \includegraphics[width=0.19\textwidth]{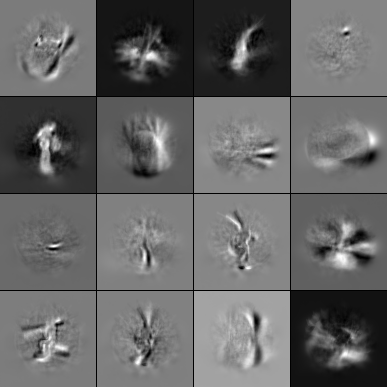}
        \hspace{0.02\textwidth}
        \includegraphics[width=0.19\textwidth]{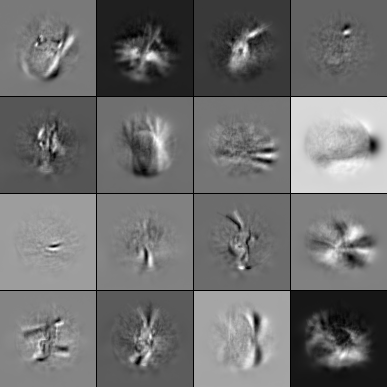}
        \label{fig:norb_filters}
    }
    \caption{HGAE first-layer filter pairs (after multi-step predictive training).}
    \label{fig:filters}
\end{figure*}

\begin{figure}[ht]
\vskip 0.2in
\begin{center}
\centerline{\includegraphics[width=0.6\columnwidth]{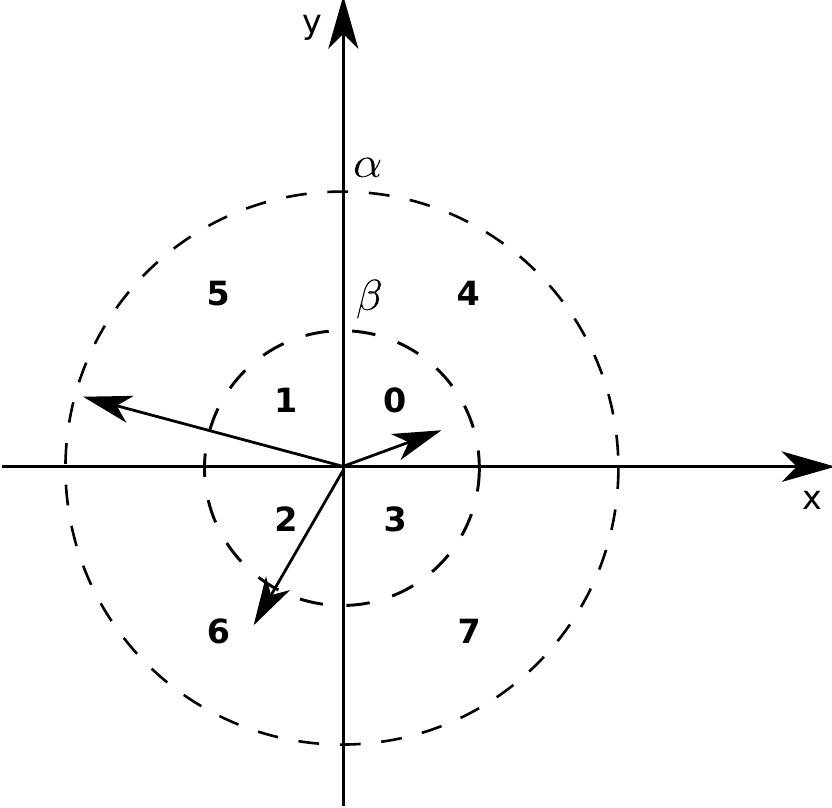}}
\caption{Discretization of shift vectors: The 2D plane is divided into the four 
quadrants and then a magnitude threshold $\beta$ is chosen, such that the distribution
of samples into the 8 shown bins (numbered $0-7$) is uniform. $\alpha$ denotes the maximum magnitude in 
the respective data set.}
\label{fig:shift_label_discretization}
\end{center}
\vskip -0.2in
\end{figure}

To evaluate whether predictive training of the GAE yields better representations of 
transformations than training with the reconstruction objective, a classification
experiment on videos showing artificially transformed natural images was performed.
The $13\times13$ patches were cropped from the Berkeley Segmentation data set \cite{martinftm01}.
Two data sets with videos featuring constant velocity shifts ( \textsc{ConstShift}) and 
rotations (\textsc{ConstRot}) were generated. The elements of the shift vectors for the 
$\textsc{ConstShift}$ data set were sampled uniformly from the interval $[-3,3]$ (in pixels).
The rotation angles were sampled uniformly from the interval $(-\pi, \pi)$.
Labels for the \textsc{ConstShift} data set were generated by dividing the shift vectors as 
shown in Figure \ref{fig:shift_label_discretization}. For \textsc{ConstRot} the angles were
divided into $8$ equally-sized bins. Both data sets were partitioned into a training set 
containing $100\,000$, a validation set containing $20\,000$ and a test set containing $50\,000$ 
sequences.

The numbers of filters and mapping units were chosen using a grid search. The setting with best
performance on the validation set was $256$ filters and $256$ mapping units each for both training objectives 
and both data sets.
The models were each trained for $1\,000$ epochs using stochastic gradient descent (SGD) with a 
learning rate of $0.001$ and momentum $0.9$.
For the experiment the mappings of the first two inputs were used as input to a logistic regression
classifier.
The experiment was performed multiple times on both data sets and the mean classification accuracies 
are reported in Table \ref{tab:rec_v_pred}.
In all trials the GAE trained with $1$-step predictive training achieved a higher 
accuracy than the GAE trained on the reconstruction objective. This suggests that predictive training 
is able to generate a more explicit representation of transformations, that is plagued less by image content, 
as discussed in Section~\ref{sec:pred_training}.

\begin{table}[t]
\caption{Classification accuracy of the GAE on the constant rotations 
(\textsc{ConstRot}) and constant shifts (\textsc{ConstShift}) data sets,
for reconstructive and $1$-step predictive training.}
\label{tab:rec_v_pred}
\vskip 0.15in
\begin{center}
\begin{small}
\begin{sc}
\begin{tabular}{lccr}
\hline
\abovespace\belowspace
Model & ConstRot & ConstShift \\
\hline
\abovespace
rec. training & 97.6 & 76.4 \\
pred. training  & 98.2 & 79.4 \\
\hline
\end{tabular}
\end{sc}
\end{small}
\end{center}
\vskip -0.1in
\end{table}

%% file: acc_transformations.tex
To test the hypothesis that the HGAE learns to model second-order correspondences in 
sequences, image sequences with accelerated shifts (\textsc{AccShift}) and rotations (\textsc{AccRot})
of natural image patches were generated. The patches were again cropped from the Berkeley Segmentation data 
set 
and artificially transformed with initial (angular) velocity and constant 
(angular) acceleration. The scalar angular accelerations were sampled uniformly from the interval $[-\frac{\pi}{12},\frac{\pi}{12}]$
degrees. 
The initial angular velocites were sampled from the same interval.
To get labels for classification, the angular accelerations were divided into $8$ equally sized bins.
For the accelerated shifts data set, elements of the velocity and acceleration vectors were sampled 
in the interval $[-3,3]$ (in pixels). The discretization of acceleration vectors is the same as 
for the shift vectors in \textsc{ConstShift} (see Figure \ref{fig:shift_label_discretization}).
The partition sizes are the same as for \textsc{ConstRot} and \textsc{ConstShift}.

The number of filters and mapping units was set to $512$ and $256$, respectively (after 
performing a grid search).
After pretraining the HGAE was trained with gradient descent using a learning rate of $0.0001$ and 
momentum of $0.9$, first for $400$ epochs on single-step prediction and then $500$ epochs on two-step
prediction.

After training, first- and second-layer mappings were inferred from the first three frames of the 
test sequences.
The classification accuracies using logistic regression with second-layer mappings of the HGAE ($\m[2]{1:3}$) 
as descriptor, using the individual first-layer mappings ($\m[1]{1:2}$ and $\m[1]{2:3}$), 
and using the concatenation of both first-layer mappings are reported in Table \ref{tab:hgae_acc_class} 
for both data sets (before and after predictive finetuning).

\begin{table}[t]
\caption{
Classification accuracies (\%) using different layer mappings of the HGAE before and after 2-step 
finetuning on the accelerated rotations (\textsc{AccRot}) and the accelerated shifts (\textsc{AccShift}) 
data set.
$(\m[1]{1:2},\m[1]{2:3})$ denotes the concatenation of both first-layer mappings.}
\label{tab:hgae_acc_class}
\vskip 0.15in
\begin{center}
\begin{small}
\begin{sc}
\begin{tabular}{llccr}
\hline
\abovespace\belowspace
&Descriptor & AccRot & AccShift\\
\hline
\abovespace
\multirow{4}{*}{{\rotatebox{90}{pretrained}}\hspace{3pt}} &
$\m[1]{1:2}$ & 19.4 & 20.6\\
&$\m[1]{2:3}$ & 30.9 & 33.3\\
&$(\m[1]{1:2},\m[1]{2:3})$ & 64.9 & 38.4\\
\belowspace
&$\m[2]{1:3}$& 53.7 & 63.4\\
\hline
\abovespace
\multirow{4}{*}{{\rotatebox{90}{finetuned}}\hspace{3pt}} &
$\m[1]{1:2}$ &  18.1 & 20.9\\
&$\m[1]{2:3}$ &  29.3 & 34.4\\
&$(\m[1]{1:2},\m[1]{2:3})$ & 74.0 & 42.7\\
\belowspace
&$\m[2]{1:3}$ & 74.4 & 80.6\\
\hline
\end{tabular}
\end{sc}
\end{small}
\end{center}
\vskip -0.1in
\end{table}

\begin{figure}
 \centering
  \subfigure {
    \includegraphics[width=0.48\textwidth]{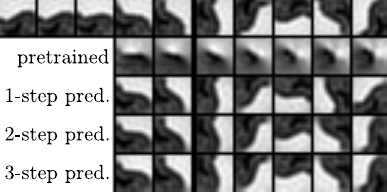}
  }\\
  \subfigure {
    \includegraphics[width=0.48\textwidth]{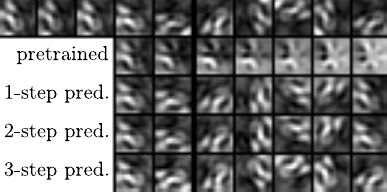}
  }
\caption{Two examples for seven prediction steps of the HGAE model on the \textsc{AccRot} data set, 
shown are from top to bottom,
groundtruth, the predictions of the model after pretraining, and after one-, two- and three-step 
predictive training.}
\label{fig:testpreds_accrot}

\end{figure} 

\begin{figure}
 \centering
  \subfigure {
    \includegraphics[width=0.28\textwidth]{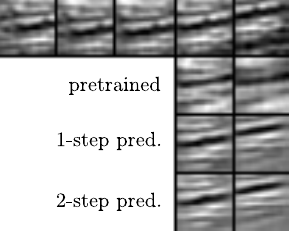}
  }
  \hspace{50pt}
  \subfigure {
    \includegraphics[width=0.28\textwidth]{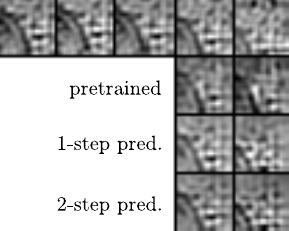}
  }
\caption{Two examples of predictions using the HGAE model on the \textsc{AccShift} data set, 
shown are from top to bottom,
groundtruth, the predictions of the model after pretraining, one- and two-step 
predictive training.}
\label{fig:testpreds_accshift}

\end{figure}

The second-layer mappings achieved a significantly higher accuracy for both data sets after predictive 
training. For the $\textsc{AccRot}$ data set, the concatenation of first-layer mappings
performed better than the second-layer mappings before finetuning, which may be because 
the angular acceleration data is based on a one-parameter transformation and is thus simpler than 
the shift acceleration data, which is based on a two-parameter transformation.
Predictive finetuning also helped improve the intermediate representation, as can be 
observed by the increase in accuracy for the concatenation of the first-layer mappings.

These results shows that the second layer of the HGAE can build a much better representation 
of the second-order relational structure in the data than the single-layer GAE model.
They further show that predictive training improves the capability of both models and 
is crucial for the two-layer model to work well. 

%% file: seq_pred.tex
In this experiment we test the capability of the models to predict previously 
unseen sequences multiple steps into the future. 
This allows us to assess to what degree modeling second order ``derivatives'' makes it 
possible to capture the temporal evolution without resorting to an explicit representation 
of a hidden state. After training, test sequences were generated by seeding the models with 
two (GAE) or three (HGAE) seed frames. Figure \ref{fig:filters} shows some of the filter pairs 
learned by the HGAE on different data sets after predictive training.

\subsubsection{Accelerated transformations}
Figures \ref{fig:testpreds_accrot} and \ref{fig:testpreds_accshift} show predictions
with the HGAE model on the data sets introduced in Section \ref{sec:acc_transformations}
after different stages of training. As can be seen in the figures, the accuracy of the 
predictions increases significantly with multi-step training.

\subsubsection{NorbVideos}
The NORBvideos data set introduced in
\cite{memisevic_aperture} contains videos of objects from the NORB dataset \cite{lecun2004learning}. 
These are objects 
divided into $5$ \emph{classes} (\emph{four-legged animals, human 
figures, airplanes, trucks} and \emph{cars}), each with $9$ \emph{instances}.
The $5$ frames of each video from the NORBvideos data show incrementally changed viewpoints 
of one of the objects.
We trained our sequence learning models on this data, using the author's original split: all videos
of objects from instances $1-8$ are in the training set and instance $9$ objects
are in the test set. This yields $109\,350$ training examples and $12\,150$ test 
examples. The frame size is $96\times96$ and the videos are $5$ frames long.
The GAE and the HGAE model were trained on the multi-step prediction task
with a learning rate of $0.0001$ and momentum $0.9$. Both models used $2000$ features and 
$1000$ mapping units (per layer). 
The test-performance of the GAE model seemed to stop improving at $2000$ features, while the
HGAE was able to make use of the additional parameters.

\begin{figure}
\begin{center}
\centering
\includegraphics[width=0.3\textwidth]{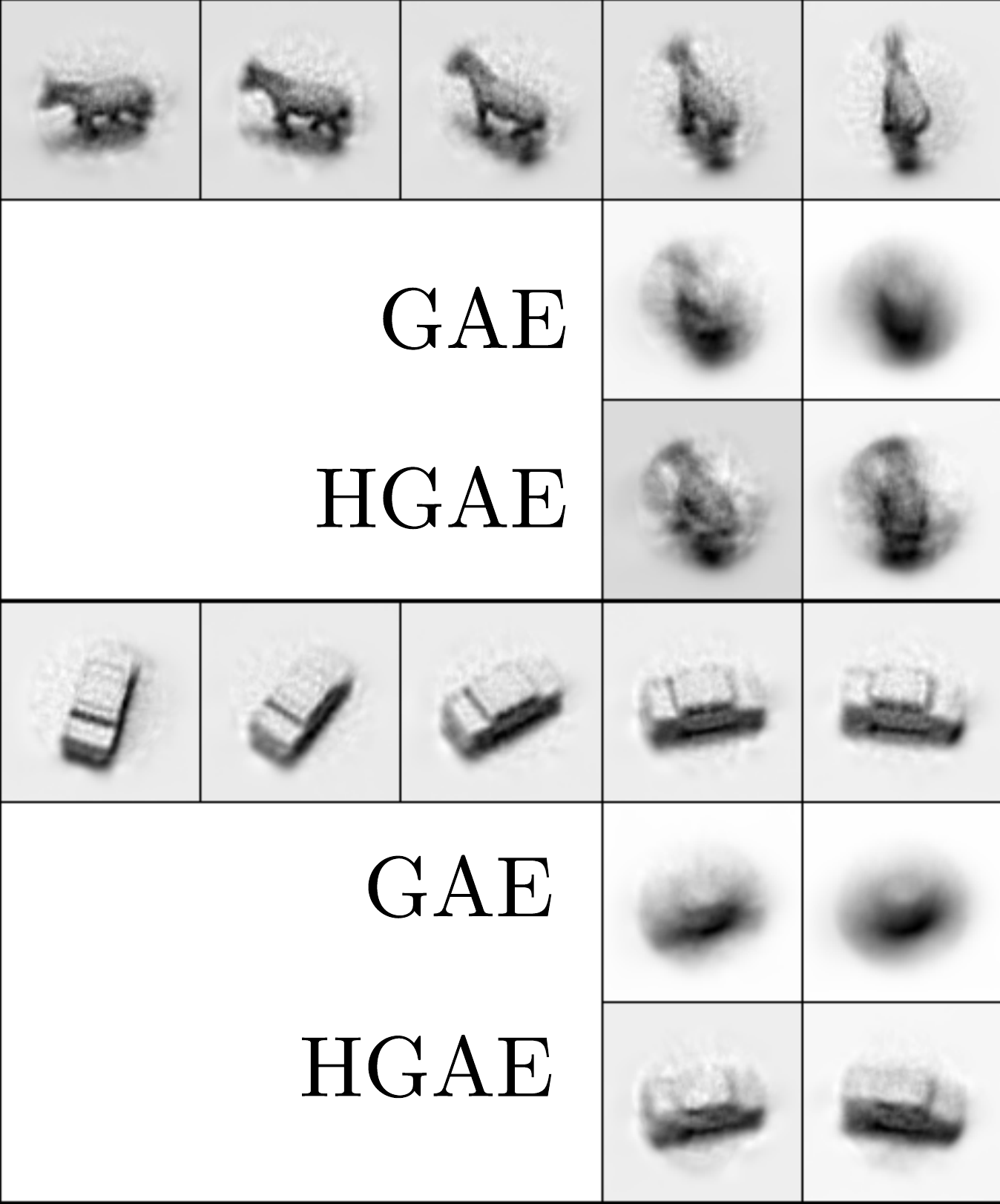}\\
\includegraphics[width=0.3\textwidth]{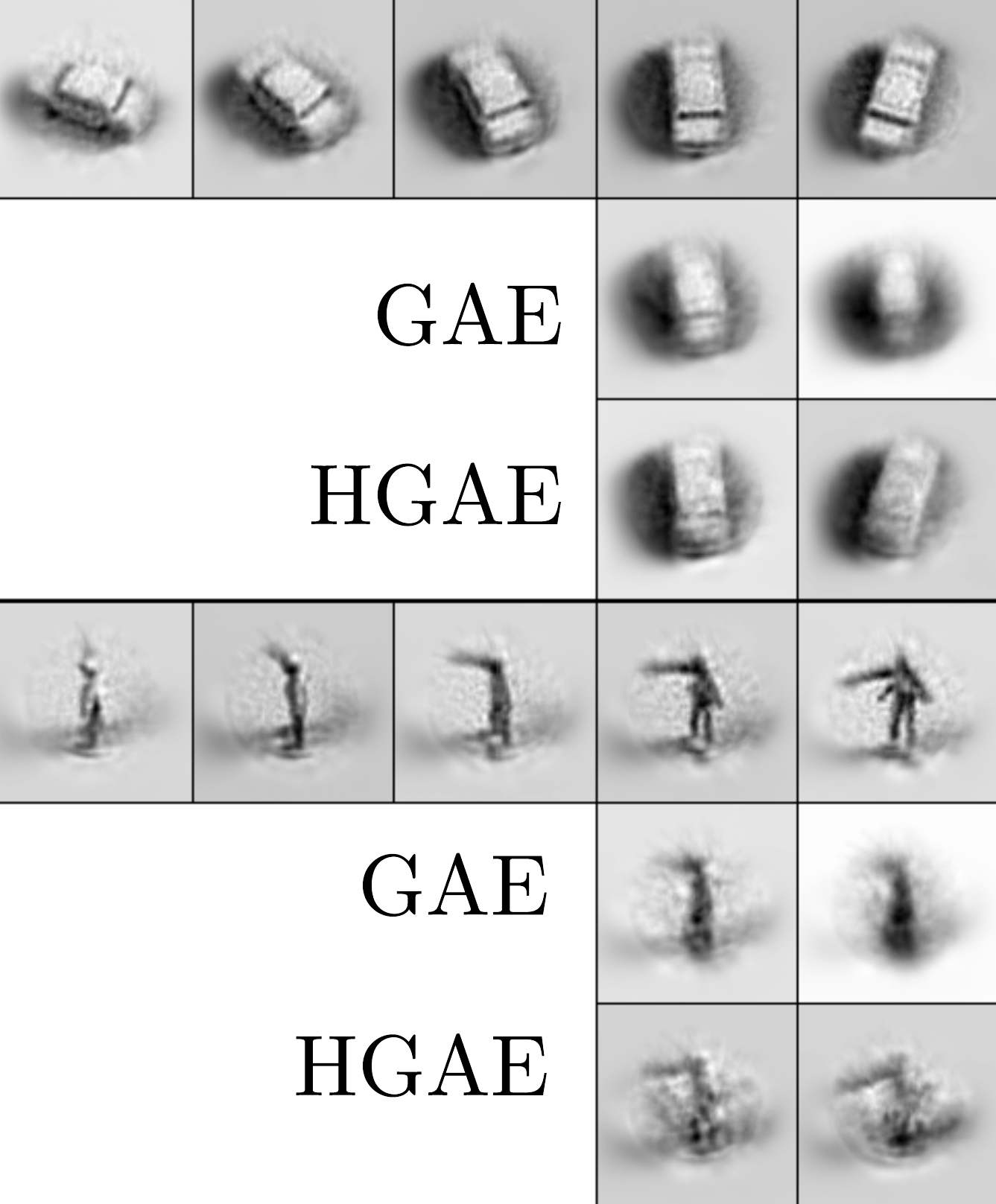}
\caption{Comparison of 10 prediction steps on the NORBvideos data set. 
    The original sequences only contain $5$ frames, providing only 2 frames of 
    ground truth for predictions.
}
\label{fig:testpreds_norb}
\end{center}
\vskip -0.2in
\end{figure}

Figure \ref{fig:testpreds_norb} shows predictions made by both models. 
The HGAE manages to generate predictions that correctly reflect the 3-D structure in the data. 
In contrast to the GAE model it is much better at extrapolating the observed transformations. 
Note that seed frames are from test data. 

Due to the large input dimensionality and the low number of training samples
a few of the filters shown in Figure \ref{fig:norb_filters} seem to be overfitting
on the training data while many others are localized Gabor-like features.

\subsubsection{Bouncing Balls}
\begin{figure}
\centering
    \includegraphics[width=0.9\columnwidth]{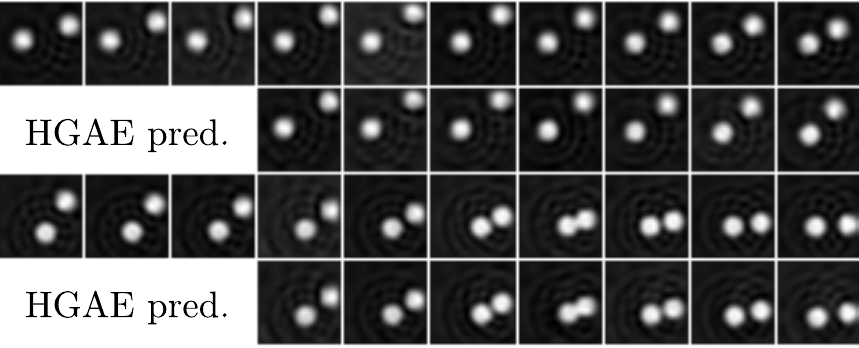}
    \caption{Seven HGAE prediction steps on two samples of the bouncing balls data set 
    after training on 3-step predictions.}
\label{fig:testpreds_bb}
\end{figure}

We also trained the HGAE on the bouncing balls data set\footnote{
    The training and test sequences were generated using the script
released with \cite{sutskever2008recurrent}.
} to see whether the HGAE
captures the highly non-linear dynamics of this data set. 
The number of features was set to $512$ and the number of mappings to $256$.
Figure \ref{fig:testpreds_bb} shows two predictions on test data. 
The predictions show that the second-order model is able to correctly 
capture reflections on the boundaries and the other balls, and makes consistent 
predictions over in some cases up to around $10-20$ frames.

%% file: discussion.tex
A major long-standing problem in sequence modeling is how to deal with long range correlations. 
It has been proposed that deep learning may help address this 
problem by finding representations that capture better the abstract, semantic content of the inputs \cite{Bengio-2009}. 
In this work we propose learning representations with the explicit goal to enable  
the prediction of the temporal evolution of the input stream multiple time steps ahead. 
Thus we seek a hidden representation that 
captures exactly those aspects of the input data which allow us to make predictions about the future. 

It is interesting to note that predictive training can also be viewed as an analogy making task \cite{memisevic2010learning}. It amounts to taking the transformation taking frame $t$ to $t+1$ and applying it to a new observation at time $t+1$ or later. The difference is that in a genuine analogy making task, the target image may be unrelated to the source image pair, whereas here target and source are related.
It would be interesting to apply the model to word representations, or language in general, as this is a domain where both, sequentially structured data and analogical relationships between data-points, play a crucial role \citep[e.g.][]{mikolov2013efficient}.

%% file: acknowledgements.tex
This work was supported by the German Federal Ministry of Education and Research (BMBF)
in project 01GQ0841 (BFNT Frankfurt), by an NSERC Discovery grant and by a Google faculty research award.